\documentclass[10pt,twocolumn,letterpaper]{article}

\usepackage{cvpr}
\usepackage{times}
\usepackage{epsfig}
\usepackage{graphicx}
\usepackage{amsmath}
\usepackage{amssymb}
\usepackage{hyperref}
\usepackage{url}
\usepackage{subfigure}
\usepackage{algorithm}
\usepackage[noend]{algpseudocode}
\usepackage{multirow}      
\usepackage{multicol}  
\graphicspath{{figure/}}
\cvprfinalcopy % *** Uncomment this line for the final submission
\def\cvprPaperID{6029} % *** Enter the CVPR Paper ID here

\begin{document}

%%%%%%%%% TITLE
\title{SQuantizer: Simultaneous Learning for \\Both Sparse and Low-precision Neural Networks}

\author{Mi Sun Park \qquad Xiaofan Xu \qquad Cormac Brick\\
Movidius, AIPG, Intel\\
{\tt\small mi.sun.park@intel.com, xu.xiaofan@intel.com, cormac.brick@intel.com}
% For a paper whose authors are all at the same institution,
% omit the following lines up until the closing ``}''.
% Additional authors and addresses can be added with ``\and'',
% just like the second author.
% To save space, use either the email address or home page, not both
%\and
%Xiaofan Xu\\
%Institution2\\
%First line of institution2 address\\
%{\tt\small secondauthor@i2.org}
}

\maketitle
%\thispagestyle{empty}

%%%%%%%%% ABSTRACT

% todo: cormac:  using the joint optimization technique 

% todo: cormac:  using the joint optimization technique 
\begin{abstract}

Deep neural networks have achieved state-of-the-art (SOTA) accuracies in a wide range of computer vision, speech recognition, and machine translation tasks. However the limits of memory bandwidth and computational power constrain the range of devices capable of deploying these modern networks. To address this problem, we propose SQuantizer, a new training method that jointly optimizes for both sparse and low-precision neural networks while maintaining high accuracy and providing a high compression rate.  This approach brings sparsification and low-bit quantization into a single training pass, employing these techniques in an order demonstrated to be optimal. 

Our method achieves SOTA accuracies using 4-bit and 2-bit precision for ResNet18, MobileNet-v2 and ResNet50, even with high degree of sparsity. 
The compression rates of 18$\times$ for ResNet18 and 17$\times$ for ResNet50, and 9$\times$ for MobileNet-v2 are obtained when SQuantizing\footnote{SQuantizing: joint optimization of Sparsification and low-precision Quantization.} both weights and activations within 1\% and 2\% loss in accuracy for ResNets and MobileNet-v2 respectively.  An extension of these techniques to object detection also demonstrates high accuracy on YOLO-v3.
Additionally, our method allows for fast single pass training, which is important for rapid prototyping and neural architecture search techniques.

Finally extensive results from this simultaneous training approach allows  us to draw some useful insights into the relative merits of sparsity and quantization.

%Furthermore, our extension effort on YOLO-v2 for object detection task demonstrates the general applicability of the proposed SQuantizer. %, which gives 17$\times$ compression rate within 1\% loss in accuracy when SQuantizing weights.

\end{abstract}

%%%%%%%%% BODY TEXT
\section{Introduction}

%As almost all previous approaches of both sparsification and quantization only look at the magnitude of gradient, we believe that we are opening a new door for this field
%~\cite{he2016deep}~\cite{huang2017densely}~\cite{sandler2018mobilenetv2} 

High-performing deep neural networks~\cite{he2016deep, huang2017densely, sandler2018mobilenetv2} consist of tens or hundreds of layers and have millions of parameters requiring billions of float point operations (FLOPs). Despite the popularity and superior performance, those high demands of memory and computational power make it difficult to deploy on resource-constrained edge devices for real-time AI applications like intelligent cameras, drones, autonomous driving, and augmented and virtual reality (AR/VR) in retail. %, and smart healthcare. 
To overcome this limitation, academia and industry have investigated network compression and acceleration in various directions towards reducing complexity of networks, and made tremendous progresses in this area. It includes network pruning ~\cite{Han2016ICLR, guo2016dynamic, xu2018hybrid}, network quantization \cite{mishra2018apprentice, zhuang2018towards, Choi2018BridgingTA}, low-rank approximation \cite{Tai2015, 8099498}, efficient architecture design \cite{sandler2018mobilenetv2, ShuffleNet}, neural architecture search \cite{NAS, Gordon2017MorphNetF} and hardware accelerators \cite{han2016eie, Parashar}.

In this work, we focus on combining the two popular network compression techniques of sparsification and quantization into a single joint optimization training. 
In addition to reduce the training time by half, our method maximizes compression which is higher than applying either the technique alone with SOTA accuracy, and further enables significant compute acceleration on deep neural networks (DNNs) hardware accelerators like Tensilica IP~\footnote{\url{https://ip.cadence.com/ai}} and Samsung sparsity-aware Neural Porcessing Unit~\cite{Song2019Samsung}. This is a key enabler of fast and energy efficient inference.
As the relative merits of sparsity and quantization are discussed in Section \ref{sec:discussion}, either the method alone can not provide optimal performance. A few previous works ~\cite{Han2016ICLR, mathew2017sparse} applied both the techniques one after the other to show that pruning and 8-bit weight only quantization can work together to achieve higher compression. However, applying one after the other not only requires two-stage training, but also makes it difficult to quantize with lower precision after pruning, due to the lack of understanding the impact of pruning weights on quantization, and vice versa.
We therefore aim for more efficient network training process with both sparse low-precision weights and sparse low-precision activations.

The main contributions of this work are summarized as below: (1) we propose a new training method to enable simultaneous learning for sparse and low-precision neural networks that sparsify and low-bit quantize both weights and activations. (2) we analyze the order effect of sparsification and quantization when trained together for optimal performance. (3) we extensively evaluate the effectiveness of our approach on ImageNet classification task using ResNet18, ResNet50 and MobileNet-v2, and also extend to object detection task using YOLO-v3. The comparision to prior quantization works shows that our method outperforms across networks, even with additional high degree of sparsity for further reduction in model size, memory bandwidth, energy and compute.

\begin{figure*}[!htbp]
\begin{center}
\includegraphics[trim={20cm 8cm 13cm 10cm},clip,width=1.0\linewidth]{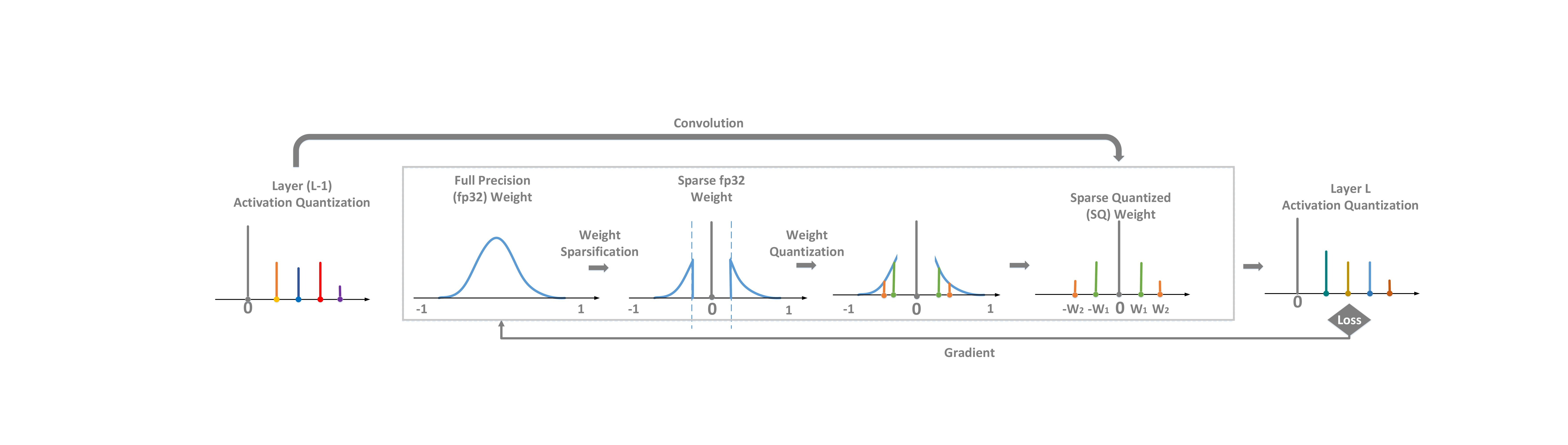}
\end{center}
\caption{Overview of our proposed SQuantizer procedure for sparse and low-precision networks. The gray box shows the weight sparsification and quantization (SQuantization) steps in layer $l$  in forward pass. The resulting sparse and low-bit quantized weights in layer $l$ are then convolved with low-bit quantized output activations of the previous layer $l-1$. In backward pass, the full-precision weights are updated with the gradients of SQuantized weights and activations at each iteration of training. Best view in color.}
%\caption{Overview of training a model using SQuantizer. The box contains the step for proposed SQuantizer for the weights in layer $l$. The sparse and quantized weights are convolved with the quantized activation from layer $(l-1)$ and produce quantized activation for layer $l$ after ReLU. The loss is being calculated after the quantized output and update the full precision weights in the backward pass.}
\label{fig:overview}
\end{figure*}

\section{Related Work}
% not multi-bit quantization
\textbf{Network Pruning:} With a goal of easy deployment on embedded systems with limited hardware resources, substantial efforts have been made to reduce network size by pruning redundant connections or channels from pre-trained models, and fine-tune to recover original accuracy.  While many of the related approaches differ in the method of determining the importance of weights or channels, the basic goal of removing unimportant weights or channels from original dense models to generate sparse or compact models remains the same. 
Specifically, fine-grained weight pruning (aka sparsification) ~\cite{8100126, Han2016ICLR, guo2016dynamic, srinivas2015data, gale2019state} seeks to remove individual connections, while coarse-grained pruning ~\cite{he2017channel, ThiNet_ICCV17, Li2016PruningFF, Molchanov2016PruningCN, ye2018rethinking} seeks to prune entire rows/columns, channels or even filters. 

%For fine-grained weight pruning

Deep compression \cite{Han2016ICLR} introduced three-stage training of pruning, trained quantization and Huffman coding and showed that weight pruning and weight quantization can work together for higher compression. %show 35$\times$ weight compression rate for AlexNet \cite{Krizhevsky}. 
Dynamic network surgery \cite{guo2016dynamic} presented an idea of connection splicing to restore incorrectly pruned weights from the previous step. 
Energy-aware pruning \cite{8100126} proposed layer-by-layer energy-aware weight pruning for energy-efficient CNNs that directly uses the energy consumption of a CNN to guide the pruning process. %In general, layer-by-layer pruning gives better performance than globally pruning, at the expense of increased training time.

ThiNet \cite{ThiNet_ICCV17} presented a filter level pruning method for modern networks with consideration of special structures like residual blocks in ResNets.
AMC~\cite{he2018amc} introduced AutoML for structured pruning using reinforcement learning to automatically find out the effective sparsity for each layer and remove input channels accordingly. Generally speaking, coarse-grained pruning results in more regular sparsity patterns, making it easier for deployment on existing hardware for speedup but more challenging to maintain original accuracy. On the other hand, fine-grained weight pruning results in irregular sparsity patterns, requiring sparse matrix support but relatively easier to achieve higher sparsity.

\textbf{Network Quantization:} 
Network quantization is another popular compression technique to reduce the number of bits required to represent each weight (or activation value) in a network. Post-training quantization and training with quantization are two common approaches in this area. Post-training quantization is to quantize weights to 8-bit or higher precision from a pre-trained full-precision network with and without fine-tuning. 
Google Tensorflow Lite~\footnote{\url{https://www.tensorflow.org/lite/performance/post_training_quantization}} and Nvidia TensorRT~\footnote{\url{https://developer.nvidia.com/tensorrt}} support this functionality by importing a pre-trained full-precision model and converting to 8-bit quantized model. %Although it allows fast conversion to lower bits but hard to maintain original accuracy.

Significant progress has recently been made on training with quantization approach for low-precision networks \cite{zhou2016dorefa, zhu2016trained, zhou2017incremental, Choi2018BridgingTA, zhou2018explicit}. 
DoReFa-Net \cite{zhou2016dorefa} presented a method to train CNNs that have low-precision weights and activations using low-precision parameter gradients. TTQ \cite{zhu2016trained} introduced a new training method for ternary weight networks with two learnable scaling coefficients for each layer.
INQ \cite{zhou2017incremental} proposed an incremental training method of converting pre-trained full-precision network into low-precision version with weights be either powers of two or zero.
More recent researches \cite{mishra2018apprentice, zhuang2018towards, choi2018pact} have tackled the difficulty of training networks with both low-precision weights and low-precision activations. Apprentice \cite{mishra2018apprentice} used knowledge distillation technique by adding an auxiliary network to guide for improving the performance of low-precision networks. Low-bit CNN \cite{zhuang2018towards} presented three practical methods of two-stage optimization, progressive quantization and guided training with full-precision network. PACT \cite{choi2018pact} proposed a new activation quantization function with a learnable parameter for clipping activations for each layer.

\section{Our Method}

In this section, we first introduce our SQuantizer for simultaneous learning for sparse and low-precision neural networks, and analyze the order effect of sparsification and quantization techniques when trained together. Furthermore, we elaborate on the details of our sparsification and quantization methods.

\subsection{Learning both sparse and low-precision values}

Our proposed SQuantizer method is illustrated in Figure \ref{fig:overview}. In each forward pass of training, we first sparsify full-precision weights based on a layer-wise threshold that is computed from the statistics of the full-precision weights in each layer. %, using binary weight mask. 
Then, we quantize the non-zero elements of the sparsified weights with min-max uniform quantization function (i.e. The minimum and maximum values of the non-zero weights). 
In case of activation, prior sparsification is not necessary, since output activations are already sparse due to the non-linearity of ReLU-like function. In general, ReLU activation function can result in about 50\% sparsity.
Therefore, we only quantize the output activations after batch normalization and non-linearity, which is also the input activations to the following convolutional layer. 

In backward pass, we update the full-precision dense weights with the gradients of the sparse and low-bit quantized weights and activations. The gradients calculation for the non-differential quantization function is approximated with the straight-through estimator (STE) \cite{Bengio_estimatingor}. Our method allows dynamic sparsification assignment and quantization values by leveraging the statistics of the full-precision dense weights in each iteration of training.

After training, we discard the full-precision weights and only keep the sparse and low-bit quantized weights to deploy on resource-constrained edge devices. 
It is worth noting that, in case of activation quantization, we still need to perform on-the-fly quantization on output activations at inference time because it is dynamic depending on input images, unlike weights.

\subsection{Analysis of the order effect}

We analyze the order effect of the two compression (sparsification and quantization) techniques when applied together. Fundamentally, we want to find the optimal order that possibly leads to better performance. The two candidates are quantization followed by sparsification (\textit{S on Q}), and sparsification followed by quantization (\textit{Q on S}). %sparsification on quantized weights (\textit{S on Q}) and (2) quantization on sparse weights (\textit{Q on S}). 

\begin{figure}[!htbp]
\vspace{-4mm}
\begin{center}
\includegraphics[width=\linewidth]{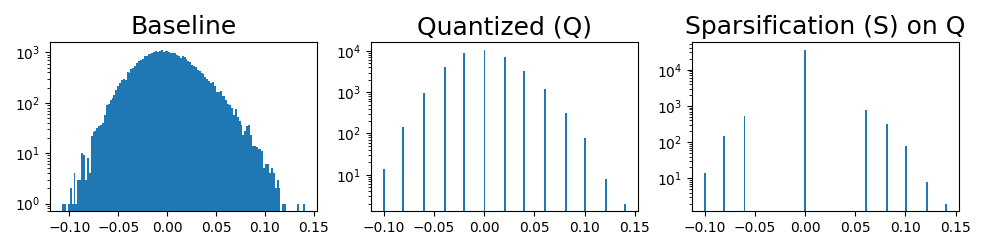}
\includegraphics[width=\linewidth]{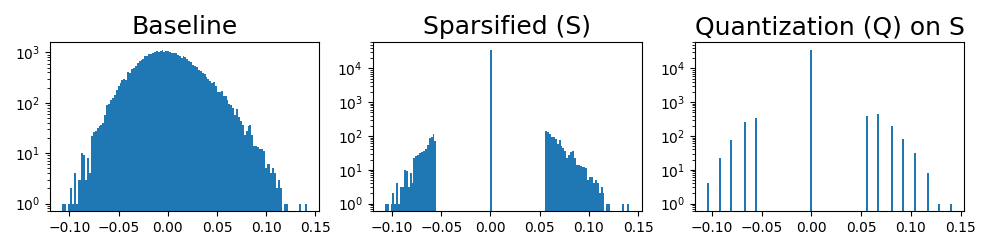}
\end{center}
\caption{Weight histogram of ResNet56 layer3.5.conv2 layer on Cifar-10 using two different orders (\textit{S on Q} and \textit{Q on S}). X-axis are weight values and Y-axis are frequency at log-scale.}
\label{fig:SonQ_resne56}
\vspace{-4mm}
\end{figure}

Figure \ref{fig:SonQ_resne56} shows the weight histograms of layer3.5.conv2 layer (the last 3$\times$3 convolutional layer in ResNet56) before and after applying the two techniques based on the two different orders. 
The top \textit{S on Q} subfigure shows three histograms of full-precision baseline, 4-bit quantized weights, and sparse and 4-bit quantized weights (after sparsifying the quantized weights) respectively. 
The bottom \textit{Q on S} subfigure represents the histograms of baseline, sparse weights, and sparse and 4-bit quantized weights (after quantizing the sparsified weights) respectively. 

It is observed from the histograms of \textit{S on Q}, we don't fully utilize all quantization levels. Although we can use up to $2^{4}$ levels for 4-bit quantization, we end up using fewer levels due to the subsequent sparsification. 
The higher sparsity, the more number of quantization levels will be underutilized. 
Noting that this phenomenon heavily depends on sparsification methods. With random sparsification, you may still utilize all the levels and in this case, the order doesn't matter. However, magnitude-based methods \cite{han2015learning, guo2016dynamic, xu2018hybrid, gale2019state} are commonly used and work better in practice. 
In fact, with \textit{Q on S}, magnitude-based methods reduce the dynamic range of weights, thus reduce quantization error with finer quantization steps. 
%Therefore, we believe \textit{Q on S} is more effective than \textit{S on Q}.  

\begin{table}[htb]
  \center
  \caption{Validation accuracy of sparse and 4-bit quantized ResNet56 on Cifar-10 using two different orders (\textit{S on Q} vs \textit{Q on S}). W and A represent weight and activation respectively. \label{tab:resnet56}}
   \resizebox{\columnwidth}{!}{%
\begin{tabular}{ccccc}
\hline
\multirow{2}{*}{} & \multicolumn{2}{c}{\textit{S on Q}} & \multicolumn{2}{c}{\textit{Q on S}} \\
\cline{2-3}
\cline{4-5}
    & Top1 (\%) & Sparsity (\%) &  Top1 (\%) & Sparsity (\%) \\
\hline
baseline &  93.37 & 0 & 93.37 & 0 \\
\hline
sparse (4W, 32A) & 93.42 & 57 & \textbf{93.45} & 57 \\ %0.0f, 58
\hline
sparse (4W, 32A) & 93.34 & 73 & \textbf{93.40} & 73\\ %0.4f
\hline
sparse (4W, 4A) & 92.88 & 57 & \textbf{92.94} & 57 \\ %0.0f
\hline
\end{tabular} \vspace{-0.15in}
}
\end{table}

We also conduct experiments to verify our analysis with ResNet56 on Cifar-10, as shown in Table \ref{tab:resnet56}. As expected, \textit{Q on S} consistently outperforms \textit{S on Q} in all three experiments. 
Moreover, our sparse and 4-bit quantized models give slightly better accuracy than the baseline, possibly because SQuantization acts as additional regularization which helps prevent overfitting. 
In summary, we believe that \textit{Q on S} is more effective than \textit{S on Q}, therefore we use \textit{Q on S} in our SQuantizer.

\iffalse
In this work we analysis the weight distribution for both quantization on sparse weights (Q on S) and sparsification on quantized weights (S on Q). Figure\ref{fig:sq+qs} shows the training flow for both methods. Different from the Q on S explained in Section\ref{sec:networkspar} and\ref{sec:networkquantization}, S on Q perform quantization on the full precision dense weights in the model first. Depends on the choice of $k$ bits, the weights in the $n$ iteration are quantized based on Equation\ref{eq:weight_quat} by substituted $SW_{l}^{n}(i,j)$ with $W_{l}^{n}(i,j)$ gives result as $QW_{l}^{n}(i,j)$. Afterwards, Equation\ref{eq:pruneweights} is applied on the $QW_{l}^{n}(i,j)$ and finally $SQW_{l}^{n}(i,j)$ can be obtained again. It is worth to taking note that the weights from both methods will be sparse and quantized.  

\begin{figure}
	\vspace{-3mm}
	\begin{center}
    \subfigure[]{
  \includegraphics[width=0.1\linewidth,height=40mm]{sq.png}
  \label{fig:sq}}
  \subfigure[]{  
  \includegraphics[width=0.1\linewidth,height=40mm]{qs.png}
  \label{fig:qs}}
  \end{center}
  \caption{Training flow for (a) Quantization on sparse weights (Q on S), (b) Sparsification on quantized weights (S on Q)}
  \label{fig:sq+qs}
\end{figure}
\fi

\subsection{SQuantizer in details}
\subsubsection{Sparsification}
\label{sec:networkspar}

As shown in Figure \ref{fig:overview}, we first apply statistic-aware sparsification to prune connections in each layer by removing (zeroing out) the connections with absolute weight values lower than a layer-wise threshold.
The basic idea of statistic-aware sparsification is to compute a layer-wise weight threshold based on the current statistical distribution of the full-precision dense weights in each layer, and mask out weights that are less than the threshold in each forward pass. In backward pass, we also mask out the gradients of the sparsified weights with the same mask.

We use layer-wise binary $mask_{l}^{n}$ (same size as weight $W_{l}^{n}$) for $l^{th}$ layer at $n^{th}$ iteration in Equation \ref{eq:mask} and Equation \ref{eq:threshold}. Similar to \cite{guo2016dynamic, xu2018hybrid}, this binary mask is dynamically updated based on a layer-wise threshold and sparsity controlling factor $\sigma$ (same for all layers). The mean and one standard deviation (std) of the full-precision dense weights in each layer are calculated to be a layer-wise threshold. 
This allows previously masked out weights back if it becomes more important (i.e. $ |W_{l}^{n}(i,j)| > t_{l}^{n}$). 
It should be noted that our approach doesn't need layer-by-layer pruning \cite{han2015learning, 8100126}. It globally prunes all layers with layer-wise thresholds considering different distribution of weights in each layer. 
Our experiment shows that our statistics-aware method performs better than globally pruning all layers with the same sparsity level, and is comparable to layer-by-layer pruning with much less training epochs. 
%Moreover, in order to perform both sparsification and quantization simultaneously, we prefer global pruning method over layer-by-layer method due to 

\begin{equation}
mask_{l}^{n}(i,j) = \begin{cases}
0 & \text{ if } \left | W_{l}^{n}(i,j) \right| < t_{l}^{n} \\ 
1 & \text{ if } \left | W_{l}^{n}(i,j) \right| > t_{l}^{n}
\end{cases}
\label{eq:mask}
\end{equation}

\begin{equation}
t_{l}^{n} = mean(|W_{l}^{n}|) + std(|W_{l}^{n}|)\times \sigma
\label{eq:threshold}
\end{equation}

\begin{figure}[!htbp]
\vspace{-4mm}
\begin{center}
\includegraphics[trim={2.5cm 0.1 2.5cm 0.1},width=\linewidth]{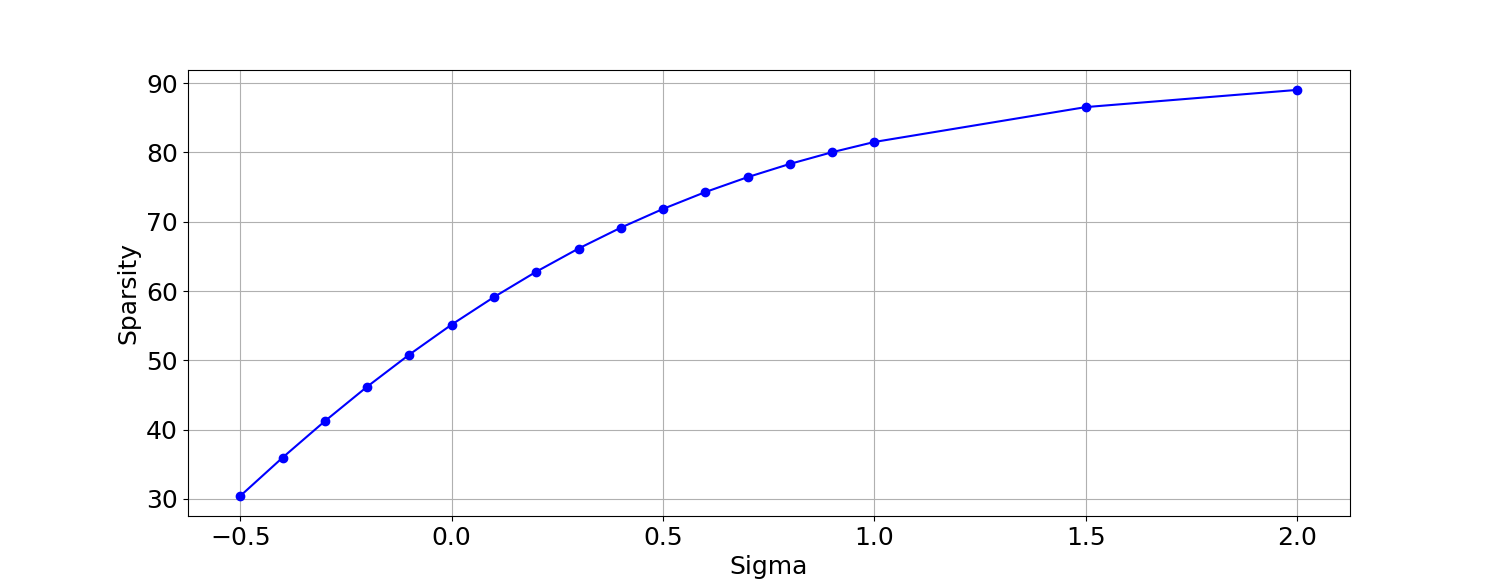}
\end{center}
\vspace{-4mm}
\caption{$\sigma$ vs sparsity of sparse and 4-bit ResNet50. X-axis is sigma ($\sigma$) and Y-axis is sparsity level (\%).}
\label{fig:sigma}
\vspace{-4mm}
\end{figure}

Sparsity controlling factor $\sigma$ is a hyper-parameter in this statistic-aware pruning method. Unlike explicit level of target sparsity (i.e. prune 50\% of all layers), $\sigma$ is implicitly determining sparsity level. %Thus, determining the proper value of $\sigma$ for target sparsity is important. 
To understand the effect of $\sigma$ on accuracy and sparsity level, we experiment for 4-bit and 2-bit ResNet50 on ImageNet, shown in Table \ref{tab:variousFactors_resnet50} and Figure \ref{fig:sigma}. 
As expected, the higher $\sigma$, the more sparsity we can get with a slight decrease in accuracy. 
We can achieve up to 30$\times$ compression rate for sparse and 4-bit model within 1\% drop in accuracy, while achieving up to 42$\times$ compression rate for sparse and 2-bit model within 2\% drop in accuracy.
%This is especially important for resource-constrained edge devices. 

\begin{table}[htb]
\scriptsize
  \center
  \caption{Effect of various $\sigma$ on accuracy and sparsity of ResNet50 on ImageNet. W represents weight. \label{tab:variousFactors_resnet50}}
     \resizebox{\columnwidth}{!}{%
\begin{tabular}{ccccccc}
\hline
\multicolumn{1}{c}{\multirow{1}{*}{}} &  & \multicolumn{2}{c}{Accuracy (\%)} & \multirow{2}{*}{Sparsity (\%)} & \multirow{2}{*}{\#Params (M)} & \multirow{1}{*}{Compression}\\
\cline{3-4}
 \multicolumn{1}{c}{} &$\sigma$ & Top1  & Top5 & & & Rate\\
\hline
\multicolumn{1}{c}{baseline} & & 76.3 &  93.0 &  0 & 25.5 & -\\
\hline
\multirow{4}{*}{\textbf{4W}} &  0.0 & 76.0 & 92.9 & 55 & 11.5 & 17$\times$ \\
%\cline{2-6}
&  0.2 & 76.0 & 92.6 & 63 & 9.5 & 21$\times$\\
%\cline{2-6}
& 0.4 & 75.4 & 92.5 & 69 & 7.9 & 25$\times$\\
%\cline{2-6}
&  0.6 & \textbf{75.3} & 92.3 & \textbf{74} & 6.6 & \textbf{30$\times$}\\
\hline
\multirow{3}{*}{\textbf{2W}} &  0.0 & 74.8 & 92.2 & 56 & 11.3 & 36$\times$\\
%\cline{2-6}
& 0.1 & 74.6 & 92.1 & 59 & 10.4 & 39$\times$\\
%\cline{2-6}
& 0.2 & \textbf{74.5} & 92.0 & \textbf{63} & 9.5 & \textbf{42$\times$} \\
\hline
\end{tabular} \vspace{-0.15in}
}
\end{table}

\subsubsection{Quantization}
\label{sec:networkquantization} 

After masking out relatively less important weights, we quantize the non-zero elements of sparsified weights with low-bitwidth $k$, as shown in Figure \ref{fig:overview}. %In case of activation, we quantize output activations (after convolution, batch normalization, and non-linearity) that are also input activations in the next conv layer. 
For weight quantization, we use min-max uniform quantization function without clipping to [-1, 1]. Our min is the previously determined layer-wise pruning threshold $t_{l}^{n}$, while the max is the maximum value of the sparse weights $sparseW_{l}^{n}$ in $l^{th}$ layer at $n^{th}$ iteration of training. %Based on Equation \ref{eq:minmax}, ~\ref{eq:w_r}, ~\ref{eq:w_q}, and ~\ref{eq:w_sq}, 
Based on Equation \ref{eq:minmax} to Equation \ref{eq:w_sq}, we quantize a full-precision non-zero element of sparse weight $sparseW_{l}^{n}(i,j)$ to $k$-bit $w_{sq}$.

\begin{equation}
\begin{matrix}
max = max(|sparseW_{l}^{n}|),\quad  min = t_{l}^{n}
\end{matrix}
\label{eq:minmax}
\end{equation}

\begin{equation}
\begin{matrix}
w_{s} = \frac{|sparseW_{l}^{n}(i,j)| - min}{max - min}
\end{matrix}
\label{eq:w_r}
\end{equation}

\begin{equation}
\begin{matrix}
w_{q} = \frac{1}{2^{k-1}-1}{round((2^{k-1}-1){w_{s}})}
\end{matrix}
\label{eq:w_q}
\end{equation}

\begin{equation}
\begin{matrix}
w_{sq} = sign(sparseW_{l}^{n}(i,j))\big({w_{q}(max - min) + min}\big)
\end{matrix}
\label{eq:w_sq}
\end{equation}

In backward pass, in order to back-propagate the non-differentiable quantization functions, we adopt the straight-through estimator (STE) approach \cite{Bengio_estimatingor}. To be more specific, we approximate the partial gradient $\frac{\partial w_{q}}{\partial w_{s}}$  and $\frac{\partial w_{sq}}{\partial w_{q}}$ with an identity mapping, to be $\frac{\partial w_{q}}{\partial w_{s}} \approx 1$ and $\frac{\partial w_{sq}}{\partial w_{q}} \approx 1$ respectively. In other words, we use the identity mapping to simply pass through the gradient untouched to overcome the problem of the gradient of round() and sign() operations being zero almost everywhere.

For activation quantization, we use parameterized clipping technique, PACT \cite{choi2018pact}. From our experiment, PACT \cite{choi2018pact} works better than static clipping to [0, 1] based activation quantization methods \cite{zhou2016dorefa, zhuang2018towards, mishra2018wrpn}. %For details, refer to PACT \cite{choi2018pact}. 
On-the-fly activation quantization at inference time has some minor costs and some significant benefits.  Costs are scaling/thresholding at the accumulator output prior to re-quantization.   
A significant benefit is the reduction in data movement (and therefore energy) required for activations, and a reduction in the storage required for heap allocated activation maps. %These reductions are particularly salient with the trend towards larger resolutions in object detection networks (YOLO), with quantized activations offering both improved ability to keep activation heap in on chip memory, and also much less bandwidth when off chip transfer is required - a significant benefit as DDR bandwidth is a critically finite resource in all deep learning compute architectures.

\textbf{Tying quantization methods to sparsity:} Depending on quantization methods, there is a case that some weights are quantized to zero giving \textit{free} sparsity. For instance, WRPN \cite{mishra2018wrpn} quantizes small weights to 0 due to clipping to [-1, 1] with implicit min of 0 and max of 1, while DoReFa-Net \cite{zhou2016dorefa} is not necessary to map the same small weights to 0, due to prior tanh transformation before quantization. %, though this method works better on dense weights from our experience. 
Mainly due to the (disconnected) bi-modal distribution of sparse weights, in Figure \ref{fig:SonQ_resne56}, we choose to use min-max quantization function to have finer quantization steps only on non-zero elements of sparse weights which in turn reduce quantization error. Noting that our method is not generating additional sparsity since the min value is always greater than zero and gets larger, as sparsity controlling $\sigma$ becomes large.

\subsubsection{Overall Algorithm}

The procedure of our proposed SQuantizer is described in Algorithm \ref{alg:overview}. %Basically, we SQuantize weights and activations in forward pass, and update full-precision dense weights with the gradients of sparse and low-bit quantized weights and activations in backward pass, at each iteration of training.
Similar to \cite{Jacob_2018_CVPR}, we introduce a $Delay$ parameter to set a delay for weight SQuantization. In other words, we start quantizing activations but defer SQuantizing weights until $Delay$ iterations to let weights stabilize at the start of training, thus converge faster. 
From our experiments, $Delay$ of one third of total training iterations works well across different types of networks, and training from scratch with $Delay$ performs better than training without $Delay$. We believe the reason is because training from scratch with $Delay$ allows enough time for weights to stabilize and fully adapt the quantized activation. 

%We believe the basic logic that allows simultaneous learning for sparse and low-precision networks is we SQuantize weights and activations in forward pass, and update full-precision weights with the gradients of sparse and low-bit quantized weights and activations in backward pass, at each iteration of training. In this way, we can consider the effect of both sparsification and quantization together and be able to update weights accordingly. 

\begin{algorithm}
\caption{SQuantization for sparse and k-bit quantized neural network}
\textbf{Input:}Training data; A random initialized full-precision model$\{W_{l}:1\leqslant l\leqslant L\}$; Weight SQuantization $Delay$; Sparsity controlling \textbf{$\sigma$}; Low-bitwidth $k$. 
\textbf{Output:} A sparse, k-bit quantized model$\{W_{sparse, quantized, l}:1\leqslant l\leqslant L\}$ 
\begin{algorithmic}[1]

\State \textbf{Step 1:} Quantize Activation:
\For {$iter = 1,...,Delay$}
%\For {$l = 1,...,Layer$}
\State Randomly sample mini-batch data
\State Quantize activation based on parameterized clipping discussed in Section \ref{sec:networkquantization}
\State Calculate feed-forward loss, and update weights $W_{l}$
%\EndFor
\EndFor
\State \textbf{end for}
\State \textbf{Step 2:} SQuantize weights and activations to $k$-bit:
\For {$iter = Delay,...,T$}
%\For {$l = 1,...,Layer$}
\State Randomly sample mini-batch data
\State With $\sigma$, calculate $W_{sparse, l}$ based on $t_{l}$ by Equation \ref{eq:threshold} layer-wisely
\State With $k, t_{l}, W_{sparse, l}$, calculate $W_{sparse, quantized, l}$ by Equation \ref{eq:minmax} to \ref{eq:w_sq} layer-wisely
\State Quantize activation based on parameterized clipping discussed in Section \ref{sec:networkquantization}
%\EndFor
\State Calculate feed-forward loss, and update weights $W_{l}$
\EndFor
\State \textbf{end for}
\end{algorithmic}
\label{alg:overview}
\end{algorithm}

\section{Experiments}
To investigate the performance of our proposed SQuantizer, we have conducted experiments on several popular CNNs for classification task using ImageNet dataset, and even extended to object detection task using COCO Dataset.
In particular, we explore the following 4 representative networks to cover a range of different architectures: 
ResNet18 with basic blocks~\cite{he2016deep}, pre-activation ResNet50 with bottleneck blocks~\cite{he2016deep}, MobileNet-v2~\cite{sandler2018mobilenetv2} with depth-wise separable, inverted residual and linear bottleneck. Futhermore, Darknet-53 for YOLO-v3 object detection \cite{redmon2018yolov3}. 

\textbf{Implementation details:} 
We implemented SQuantizer in PyTorch \cite{paszke2017automatic} and used the same hyper-parameters to train different types of networks for ImageNet classification task. During training, we randomly crop 224$\times$224 patches from an image or its horizontal flip, and normalize the input with per-channel mean and std of ImageNet with no additional data augmentation. We use Nesterov SGD with momentum of 0.9 and weight decay of $1e^{-4}$, and learning rate starting from 0.1 and divided by 10 at epochs 30, 60, 85, 100. Batch size of 256 is used with maximum 110 training epochs. For evaluation, we first resize an image to 256$\times$256 and use a single center crop 224$\times$224.
Same as almost all prior works \cite{zhou2016dorefa, mishra2018apprentice, zhuang2018towards, choi2018pact}, we don't compress the first convolutional (conv) and the last fully-connected (FC) layer, unless noted otherwise. It has been observed that pruning or quantizing the first and last layers degrade the accuracy dramatically, but requires relatively less computation. %Unlike \cite{Choi2018BridgingTA}, we still compress shortcut (residual) path in ResNets, unless noted otherwise.

For object detection task, we use open source PyTorch version of YOLO-v3\footnote{https://github.com/ultralytics/yolov3} with default hyper-parameters as our baseline and apply our SQuantizer on Darknet-53 backbone classifier for YOLO-v3.

\subsection{Evaluation on ImageNet}
We train and evaluate our model on ILSVRC2012 \cite{ILSVRC15} image classification dataset (a.k.a. ImageNet), which includes over 1.2M training and 50K validation images with 1,000 classes. We report experiment results of sparse 4-bit quantized ResNet18, ResNet50 and MobileNet-v2, and sparse 2-bit ResNet50, with comparison to prior works. 
%Moreover, we compare our SQuantizer with the prior quantization works with 4-bit weight and activation for ResNet18 and ResNet50, 2-bit weight and activation for ResNet50 respectively, to show the effectiveness of our proposed SQunatizer. 
It should be noted that almost all prior low-precision quantization works have not considered efficient network architecture like MobileNet. We believe that, despite the difficulty, our experiment on MobileNet-v2 will provide meaningful insights on the trade-offs between accuracy and model size, especially for severely resource-constrained edge devices. 

%Interestingly, it is noticeable from literature reviews in network compression, the most popular networks are AlexNet and ResNets. 

\subsubsection{Sparse and 4-bit quantized Networks}

\textbf{ResNet18:} Table \ref{tab:com_resnet18} shows the comparison of Top1 validation accuracy for 4-bit (both weight and activation) quantized ResNet18 with prior works. Noting that the DoReFa-Net \cite{zhou2016dorefa} number is cited from PACT \cite{choi2018pact}, and in the case of \textit{our \textbf{-}Quantizer}\footnote{Our \textbf{-}Quantizer is used to represent the proposed SQuantizer with sparsification disabled.}, we set $t_{l}^{n}$ to 0 in Equation \ref{eq:minmax} to disable sparsification prior to quantization. Our SQuantizer outperforms the prior works, even with 57\% of sparsity. Additionally, our \textbf{-}Quantizer achieves the state-of-the-art accuracy of 4-bit ResNet18 without sparsification.

\begin{table}[htb]
  \center
  \caption{Comparison of validation accuracy of 4-bit quantized ResNet18 on ImageNet. W and A represent weight and activation respectively. \label{tab:com_resnet18}}
     \resizebox{\columnwidth}{!}{%
\begin{tabular}{cccccc}
\hline
\multirow{2}{*}{} & \multicolumn{1}{c}{DoReFa-Net$^*$\cite{zhou2016dorefa}}  & \multicolumn{1}{c}{PACT\cite{choi2018pact}}  & \multicolumn{1}{c}{Our \textbf{-}Quantizer} & \multicolumn{2}{c}{Our SQuantizer}\\
\cline{2-6}
    & Top1  & Top1 & Top1 & Top1 & Sparsity (\%) \\
\hline
baseline & 70.2  &  70.2 &  70.4 & 70.4 & 0 \\
\hline
(4W, 4A) & 68.1  & 69.2 &  \textbf{69.7} & \textbf{69.4} & \textbf{57}\\
\hline
\end{tabular} 
}\vspace{-2mm}
\end{table}

In Table \ref{tab:wSQ_ResNet18}, our \textbf{-}Quantizer is used for dense and 4-bit models (4W, -A), while our SQuantizer with $\sigma$ of 0 is used for sparse and 4-bit quantized models (sparse (4W, -A)). When SQuantizing both weights and activations, we achieve up to 18$\times$ compression rate within 1\% drop in accuracy. Due to the uncompressed last FC layer, the overall sparsity is slightly lower than the conv sparsity (57\% vs 60\%).

\begin{table}[htb]
  \center
  \caption{SQuantization on ResNet18 on ImageNet \label{tab:wSQ_ResNet18}}
  \vspace{-2mm}
\resizebox{\columnwidth}{!}{%
\begin{tabular}{ccccccc}
\hline
\multirow{2}{*}{} & \multicolumn{2}{c}{Accuracy (\%)}  & \multicolumn{2}{c}{Sparsity (\%)} & \multirow{2}{*}{\#Params (M)}& \multirow{1}{*}{Compression}\\
\cline{2-5}
 \multicolumn{1}{c}{}  & Top1  & Top5 &  Conv  & All & &Rate  \\
\hline
  baseline & 70.4 & 89.7 & 0 &  0 &11.7 & -\\
\hline
 (4W, 32A) & 70.2 & 89.4 & 0 &  0 &11.7 & 8$\times$\\
 \hline
 sparse (4W, 32A) & \textbf{69.8} & 89.2 &  60 & \textbf{57} &5.0 & \textbf{18$\times$}\\
\hline
 (4W, 4A) & 69.7 & 89.1  & 0 &  0 & 11.7 & 8$\times$\\
\hline
 sparse (4W, 4A) & \textbf{69.4} & 89.0 &  60 & \textbf{57} &5.0 & \textbf{18$\times$}\\
\hline
%\hline
%  DoReFa-Net$^*$\cite{zhou2016dorefa} baseline & 70.2 & - & - &  - &- & -\\
%\hline
% (4W, 4A) & 68.1 & - & - &  - &- & -\\
%\hline
%\hline
%  PACT baseline & 70.2 & - & - &  - &- & -\\
%\hline
% (4W, 4A) & 69.2 & - & - &  - &- & -\\
%\hline
\end{tabular} 
}\vspace{-2mm}
\end{table}

\begin{figure}[!htbp]
\vspace{-4.5mm}
\begin{center}
\includegraphics[trim={2.5cm 0.2 0.1 0.5},width=1.1\linewidth]{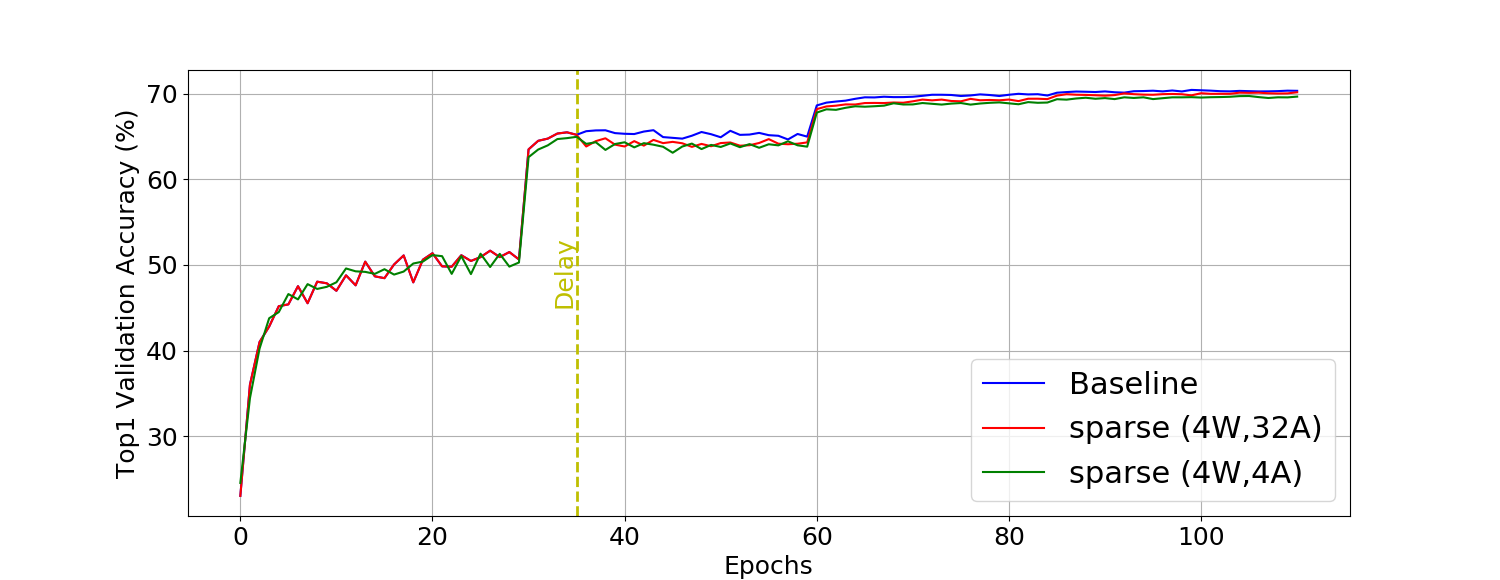}
\end{center}
\vspace{-3mm}
\caption{Top1 validation accuracy vs epochs of sparse and 4-bit ResNet18 on ImageNet. Best view in color.}
\label{fig:resnet18plot}
\vspace{-1mm}
\end{figure}

Figure \ref{fig:resnet18plot} plots Top1 validation accuracy over training epochs for baseline, sparse 4-bit weights, and sparse 4-bit weights with 4-bit activations. The $Delay$ in the plot is where step 2 starts in Algorithm \ref{alg:overview}, and accuracy curves of baseline and sparse (4W, 32A), due to the same precision of activation, are the same before $Delay$.
%And, the accuracy curves are the same for baseline and sparse 4-bit weights cases before $Delay$, since both cases use 32-bit activations. 
%To be more specific, we started SQuantization on weights at $Delay$ for sparse 4-bit weights based on the baseline model. 
Overall, the accuracy curve of our SQuantizer is similar to the baseline and shows some drop right after $Delay$ of 35 epochs, but almost recovers back at the next lowering learning rate epoch (60 epochs).

\textbf{ResNet50:} We compare our SQuantizer for 4-bit ResNet50 with many prior works in Table \ref{tab:comp_resnet50}. Noting that the DoReFa-Net \cite{zhou2016dorefa} number is cited from PACT \cite{choi2018pact} and Apprentice \cite{mishra2018apprentice} number is directly from an author of the publication. 
Our SQuantizer outperforms most of the prior works, even with sparsity. However, although PACT \cite{choi2018pact} shows better accuracy, its baseline accuracy is much higher and still loses 0.4\% accuracy after 4-bit quantization. 
Similarly, we also lose 0.4\% accuracy after 4-bit SQuantization, with 41\% sparsity ($\sigma$ of -0.3 is applied). 

%It should be noted that although we use learnable clipping method PACT \cite{choi2018pact} for our activation quantization by implementing the algorithm based on the paper, we were unable to reproduce the reported accuracy for 4-bit ResNet50, unlike ResNet18 where we outperform PACT. %We believe that it can be improved once if we can get their full implementation details for activation quantization. 

From Table \ref{tab:wSQ_ResNet50}, it is observed that our SQuantizer with 41\% sparsity gives almost as accurate as our -Quantizer. It indicates that there is an effective value for sparsity controlling $\sigma$ for each network that allows optimal sparsity with no to little loss in accuracy or even better accuracy due to additional regularization. Automatically finding out the effective value for $\sigma$ for each network will remain as our future work. In practice, we first try with $\sigma$ of 0 and then lower or raise $\sigma$ value according to the training loss in the first few epochs, and estimate the corresponding final sparsity from the intermediate model after step 1 in Algorithm \ref{alg:overview}. %We apply different configurations on the model to conduct experiments.

\begin{table}[htb]
  \center
  \caption{SQuantization on ResNet50 on ImageNet \label{tab:wSQ_ResNet50}}
  \vspace{-2mm}
\resizebox{\columnwidth}{!}{%
\begin{tabular}{ccccccc}
\hline
\multirow{2}{*}{} & \multicolumn{2}{c}{Accuracy (\%)}  & \multicolumn{2}{c}{Sparsity (\%)} & \multirow{2}{*}{\#Params (M)}& \multirow{1}{*}{Compression}\\
\cline{2-5}
 \multicolumn{1}{c}{}  & Top1  & Top5 &  Conv  & All & &Rate  \\
\hline
  baseline & 76.3 & 93.0 & 0 &  0 &25.5 &-\\
\hline
 (4W, 32A) & 76.4 & 93.0 & 0 &  0 &25.5 &8$\times$\\
 \hline
  sparse (4W, 32A) & \textbf{76.0} & 92.9 &  60 & \textbf{55} &11.5 & \textbf{17$\times$}\\
\hline
 (4W, 4A) & 76.0 & 92.7  & 0 &  0 & 25.5 &8$\times$\\
\hline
 sparse (4W, 4A) & 75.9 & 92.7 &  45 & 41 &15.0 & 13$\times$\\ %-0.3f
\hline
% sparse (4W, 4A) & 75.9 & 92.7 &  33 & 30 & & x\\ %-0.5f
%\hline
% sparse (4W, 4A) & - & - &  55 & 51 & 12.6 & x\\ %-0.1f
%\hline
 sparse (4W, 4A) & \textbf{75.5} & 92.5 &  60 & \textbf{55} &11.5 & \textbf{17$\times$} \\
\hline
\end{tabular} 
}\vspace{-2mm}
\end{table}

Table \ref{tab:wSQ_ResNet50} and Table \ref{tab:variousFactors_resnet50} show that within 1\% accuracy drop, we can achieve up to 17$\times$ compression rate for both weight and activation SQuantization, and up to 30$\times$ compression rate for weights only SQuantization. In addition, our -Quantizer for 4-bit weights performs slightly better than the baseline.

\begin{table*}[htb]
  \center
  \caption{Comparison of validation accuracy of 4-bit quantized ResNet50 on ImageNet. W and A represent weight and activation respectively.\label{tab:comp_resnet50}}
  
   \resizebox{2\columnwidth}{!}{%
\begin{tabular}{cccccccccccccc}
\hline
\multirow{2}{*}{} & \multicolumn{2}{c}{DoReFa-Net$^*$\cite{zhou2016dorefa}} & \multicolumn{2}{c}{Apprentice$^*$\cite{mishra2018apprentice}} & \multicolumn{2}{c}{Low-bit CNN\cite{zhuang2018towards}}  & \multicolumn{2}{c}{PACT\cite{choi2018pact}} & \multicolumn{2}{c}{Our \textbf{-}Quantizer} & \multicolumn{3}{c}{Our SQuantizer}  \\
\cline{2-14}
   & Top1  & Top5 & Top1  & Top5 & Top1  & Top5  & Top1  & Top5 & Top1  & Top5& Top1  & Top5& Sparsity (\%) \\
\hline
baseline & 75.6 & 92.2 & 76.2 & -  & 75.6 & 92.2 & 76.9 & 93.1 & 76.3 & 93.0 & 76.3 & 93.0 & 0\\
\hline
(4W, 4A) & 74.5& 91.5 &  75.3 & - & 75.7 & 92.0 & 76.5 & 93.2 & \textbf{76.0} & 92.7  &\textbf{75.9} & 92.7 & \textbf{41} \\
\hline
\end{tabular}
}\vspace{-2mm}
\end{table*}

\textbf{MobileNet-v2:}
MobileNets are considered to be less attractive to network compression due to its highly efficient architecture, and as such present an ardent challenge to any new network compression technique. To be more specific, MobileNet-v2 uses 3$\times$ and 7$\times$ smaller number of parameters than ResNet18 and ResNet50 respectively. 

Taking up this challenge, we applied our SQuantizer on the latest MobileNet-v2 to further reduce model size and compared with prior work, in Table \ref{tab:com_mobilentv2}  and Table \ref{tab:wSQ_mobilenet}.
Different from other networks, we sparsify the last FC layer, since the last FC layer alone uses about 36\% of total parameters, while all conv layers consume about 63\%, and the rest of 1\% is used in batch normalization. Table \ref{tab:com_mobilentv2} shows that our SQuantizer outperforms the prior work by large margin, even with sparsity. Additionally, our \textbf{-}Quantizer achieves the state-of-the-art accuracy of 4-bit MobileNet-v2 without sparsification.

\begin{table}[H]
  \center
  \caption{Comparison of validation accuracy of 4-bit quantized MobileNet-v2 on ImageNet. W and A represent weight and activation respectively. 
  \vspace{-2mm}\label{tab:com_mobilentv2}}
     \resizebox{\columnwidth}{!}{%
\begin{tabular}{ccccc}
\hline
\multirow{2}{*}{} & \multicolumn{1}{c}{QDCN\cite{Krishnamoorthi2018QuantizingDC}} & \multicolumn{1}{c}{Our \textbf{-}Quantizer} & \multicolumn{2}{c}{Our SQuantizer}\\
\cline{2-5}
    & Top1  & Top1 & Top1 & Sparsity (\%) \\
\hline
baseline & 71.9 & 72.0 & 72.0 & 0 \\
\hline
(4W, 32A) & 62.0 &71.2 & \textbf{70.7} & \textbf{25}\\
\hline
\end{tabular} 
}\vspace{-0.15in}
\end{table}

Table \ref{tab:wSQ_mobilenet} shows that within 2\% drop in accuracy, we can achieve up to 13$\times$ compression rate for weight SQuantization, and up to 9$\times$ for both weight and activation SQuantization. 
It is also observed that when applying our -Quantizer for dense 4-bit weight model, it already lost 0.8\% accuracy. For reference, ResNet18 lost 0.2\% for the same configuration. As expected, the effect of quantization is more significant on efficient network like MobileNet-v2. However, we believe this trade-offs between accuracy and compression rate can help to determine the applicability of deployment on severely resource-constrained edge devices.

\begin{figure}[!htbp]
\vspace{-4mm}
\begin{center}
\includegraphics[trim={3.5cm 0.1 0.1 0.1},width=1.1\linewidth]{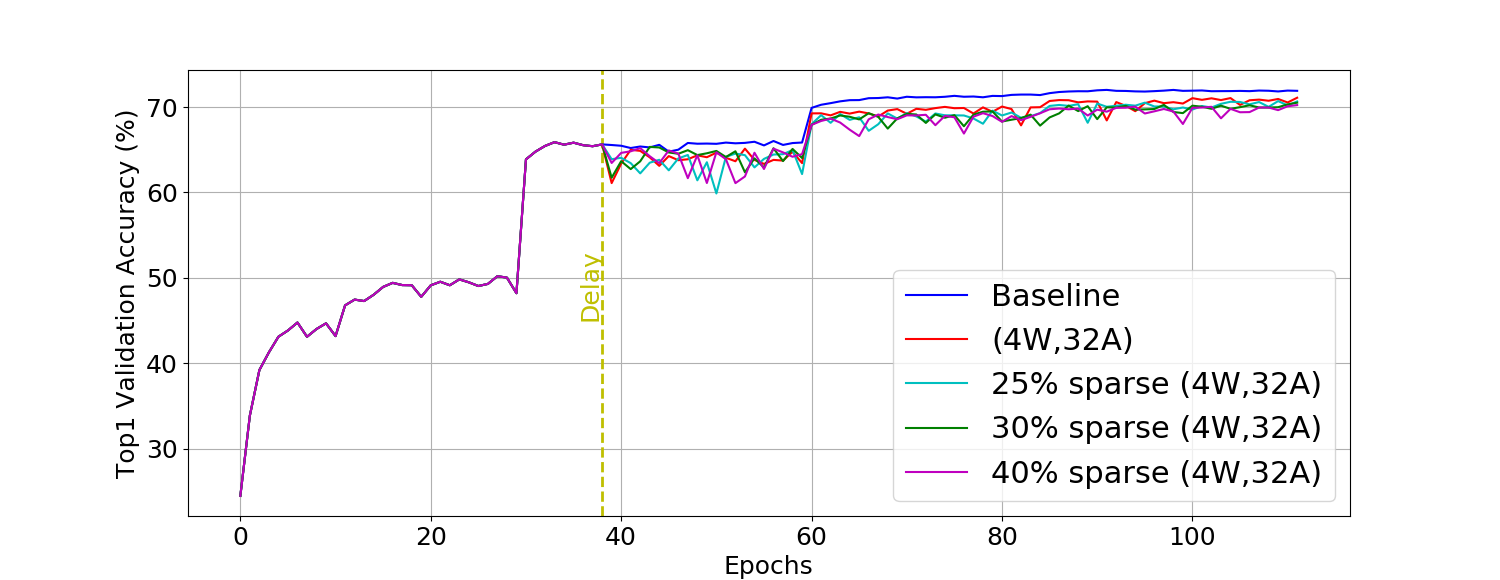}
\end{center}
\vspace{-3mm}
\caption{Top1 validation accuracy vs epochs of sparse 4-bit MobileNet-v2 on ImageNet. Best view in color.}
\label{fig:mobilenetplot}
\vspace{-4mm}
\end{figure}

Figure \ref{fig:mobilenetplot} plots Top1 validation accuracy over training epochs for baseline, dense 4-bit weights, and sparse 4-bit weights with various sparsity degree. At $Delay$ of 38 epochs, we start SQuantizing weights based on the same baseline with different configurations. It is seen that the accuracy curves of sparse 4-bit models become less stable with higher sparsity. Comparing to the dense 4-bit model, we can have up to 40\% sparsity within 1\% drop in accuracy. 
It is worth noting that training with $Delay$ allows us to reuse the same intermediate model for experiments with different configurations and shortens training time. 

\begin{table*}[htb]
  \center
  \caption{Comparison of validation accuracy of 2-bit quantized ResNet50 on ImageNet. W and A represent weight and activation respectively. The last two works  (PACT-SWAB$^*$\cite{Choi2018BridgingTA} and Our SQuantizer$^*$) use full-precision shortcut. \label{tab:2bit_resnet50}}
   \resizebox{2\columnwidth}{!}{%
\begin{tabular}{cccccccccccc||ccccc}
\hline
\multirow{2}{*}{} & \multicolumn{2}{c}{DoReFa-Net$^*$\cite{zhou2016dorefa}} & \multicolumn{2}{c}{Apprentice\cite{mishra2018apprentice}} & \multicolumn{2}{c}{Low-bit CNN\cite{zhuang2018towards}} & \multicolumn{2}{c}{PACT\cite{choi2018pact}} & \multicolumn{3}{c||}{\textbf{Our SQuantizer}} & \multicolumn{2}{c}{PACT-SWAB$^*$\cite{Choi2018BridgingTA}} & \multicolumn{3}{c}{Our SQuantizer$^*$}\\
\cline{2-17}
   & Top1  & Top5 & Top1  & Top5 & Top1  & Top5   & Top1  & Top5 & Top1  & Top5  & Sparsity (\%)& Top1  & Top5   & Top1  & Top5  & Sparsity (\%)\\
\hline
baseline & 75.6 & 92.2 & 76.2 & -  & 75.6 & 92.2 & 76.9& 93.1 & 76.3 & 93.0 & 0& 76.9 & -  & 76.3 & 93.0 & 0\\
\hline
(2W, 2A) & 67.3& 84.3 &  72.8\textit{$(8bit Act)$} & - & 70.0 & 87.5  & 72.2 & 90.5  & \textbf{73.0} & 91.0 & \textbf{40}& 74.2\textit{$(fpsc)$} & -  & \textbf{73.9}\textit{$(fpsc)$} & 91.6 & \textbf{36}\\
\hline
\end{tabular} 

}\vspace{-0.15in}
\end{table*}

\begin{table}[H]
  \center
  \caption{SQuantization on MobileNet-v2 on ImageNet \label{tab:wSQ_mobilenet}}
\resizebox{\columnwidth}{!}{%
\begin{tabular}{cccccccc}
\hline
\multirow{2}{*}{} & \multicolumn{2}{c}{Accuracy (\%)}  & \multicolumn{3}{c}{Sparsity (\%)} & \multirow{2}{*}{\#Params (M)}& \multirow{1}{*}{Compression}\\
\cline{2-6}
 \multicolumn{1}{c}{}  & Top1  & Top5 &  Conv  & FC & All & & Rate\\
\hline
  baseline & 72.0 & 90.4 & 0 & 0& 0 & 3.6 &- \\
\hline
 (4W, 32A) & 71.2 & 89.9  & 0 & 0 & 0 & 3.6 &8$\times$ \\
\hline
 sparse (4W, 32A) & 70.7 & 89.5 & 21 & 32 & 25 &  3.1 &9$\times$ \\
\hline
 sparse (4W, 32A) & 70.5 & 89.5 & 30 &32  & 30 &  2.5 &11$\times$ \\
\hline
 sparse (4W, 32A) & \textbf{70.2} & 89.3 &35  & 51 & \textbf{40} &  2.1 &\textbf{13$\times$} \\
\hline
  (4W, 4A) & 70.8 & 89.7 & 0 & 0 & 0 &  3.6 &8$\times$ \\
\hline
 sparse (4W, 4A) & \textbf{70.3} & 89.5 & 21 & 32 & \textbf{25} &  3.1 &\textbf{9$\times$} \\
\hline
%\textbf{sparse} (4W, 32A) & 70.3 & 89.2 & 35 & 36 & 35 &  2.3M\\
%\hline
% \textbf{sparse} (4W, 32A) & 70.1 & 89.3 & 41 & 42 & 41 &  2.1M\\
%\hline
%\textbf{sparse} (4W, 32A) & 69.8 & 89.1 & 47 & 48 & 47 &  1.9M\\
%\hline
\end{tabular} 
}\vspace{-2mm}
\end{table}

\iffalse
\begin{table*}[htb]
  \center
  \caption{Comparison of validation accuracy of 2-bit quantized ResNet50 on ImageNet. W and A represent weight and activation respectively. The last two works  (PACT-SWAB$^*$\cite{Choi2018BridgingTA} and Our SQuantizer$^*$) use full-precision shortcut. \label{tab:2bit_resnet50}}
   \resizebox{2\columnwidth}{!}{%
\begin{tabular}{cccccccccc||ccccc}
\hline
\multirow{2}{*}{} & \multicolumn{2}{c}{DoReFa-Net$^*$\cite{zhou2016dorefa}} & \multicolumn{2}{c}{Apprentice\cite{mishra2018apprentice}} & \multicolumn{2}{c}{Low-bit CNN\cite{zhuang2018towards}}  & \multicolumn{3}{c||}{\textbf{Our SQuantizer}} & \multicolumn{2}{c}{PACT-SWAB$^*$\cite{Choi2018BridgingTA}} & \multicolumn{3}{c}{Our SQuantizer$^*$}\\
\cline{2-15}
   & Top1  & Top5 & Top1  & Top5 & Top1  & Top5   & Top1  & Top5  & Sparsity (\%)& Top1  & Top5   & Top1  & Top5  & Sparsity (\%)\\
\hline
baseline & 75.6 & 92.2 & 76.2 & -  & 75.6 & 92.2  & 76.3 & 93.0 & -& 76.9 & -  & 76.3 & 93.0 & -\\
\hline
(2W, 2A) & 67.3& 84.3 &  72.8\textit{$(8bit Act)$} & - & 70.0 & 87.5  & \textbf{73.0} & 91.0 & \textbf{40}& 74.2\textit{$(fpsc)$} & -  & \textbf{73.9}\textit{$(fpsc)$} & 91.6 & \textbf{36}\\
\hline
\end{tabular}
}\vspace{-2mm}
\end{table*}
\fi

\subsubsection{Sparse and 2-bit quantized Network}

We SQuantize both weight and activation for 2-bit ResNet50, and compare with prior works in Table \ref{tab:2bit_resnet50}. The table contains two different comparisons, without and with full-precision shortcut (fpsc). 
%For the former, we SQuantize all the layers except the first conv and last FC layer, while 
In the case of with fpsc, we don't SQuantize input activations and weights in the shortcut (residual) path, as suggested by PACT-SWAB \cite{Choi2018BridgingTA}. 
Also note that the DoReFa-Net \cite{zhou2016dorefa} number is cited from Low-bit CNN \cite{zhuang2018towards} and Apprentice \cite{mishra2018apprentice} used 8-bit activation. %, and PACT-SWAB \cite{Choi2018BridgingTA} only reported accuracy with fpsc. 

For 2-bit quantization, we slightly modified the min and max functions in Equation \ref{eq:minmax}, for better performance. According to Equation \ref{eq:minmax}, all non-zero weights are quantized to either the smallest or largest values of non-zero weights with its own sign. 
To reduce the significant impact of largest and smallest values, we use mean and std of non-zero elements of sparse weights to determine new values of min and max. To be specific, 
we use the mean of non-zero weights as a value of new min and the sum of the mean and two std as a value of new max. Then, we clamp the absolute values of sparse weights with these new min and max.  

As shown in Table \ref{tab:2bit_resnet50}, in the case of without fpsc, our SQuantizer outperforms the prior works by large margin, even with 40\% sparsity. In the case of with fpsc, our SQuantizer with 36\% sparsity gives comparable performance considering that our baseline is (0.6\%) lower but gives (0.3\%) smaller accuracy drop compared to PACT-SWAB \cite{Choi2018BridgingTA}.
Although we used $\sigma$ of -0.3 for both the experiments, the sparsity of the latter is lower due to uncompressed shortcut path.

From Table \ref{tab:variousFactors_resnet50}, it is already seen that for sparse and 2-bit weight ResNet50, we can achieve up to 42$\times$ compression rate within 2\% accuracy drop. 
It is worth noting that our SQuantizer neither demands an auxiliary network to guide \cite{mishra2018apprentice} nor require two-stage training \cite{mathew2017sparse} to achieve state-of-the-art accuracy for sparse and low-precision models. %perform both sparsification and quantization. %Our proposed SQuantizer support simultaneous learning for both sparse and low-precision in each iteration of traning and update weights accordingly. 
%In other words, it consumes less memory for training and less training time, but 

\iffalse
More experiments are conducted on ILSVRC12 dataset to further evaluate our method. ImageNet dataset contains 1000 different categories over 1.2 million images for training and 50000 images for validation. The size for the images are various and the pre-processing stage for ImageNet dataset are based on official Pytorch~\cite{paszke2017automatic} implementation for all the models and no other manipulation are performed. The models below are tested, and PreAct ResNet50~\cite{he2016identity} is used instead of original ResNet50.

\begin{itemize}
  \item ResNet18: basic block
  \item (Preact) ResNet50: bottlenet, full pre-act
  \item MobileNet v2: depthwise separable, inverted residual and linear bottleneck
\end{itemize}  
Similarly as Cifar-10 setup, we trained all the models above from scratch based on Algorithm\ref{alg:overview}. Learning rate starts at XX and scaled XX every XX epochs. SGD is used with momentum 0.9 for optimizing, and XX for regularization. $Delay$ is set to XX for the models, and after that SQuantizer is enabled.
\fi

\iffalse
\subsubsection{lesson learned from failure}
\begin{itemize}
  \item static activation clip
  \item using KD
\end{itemize}
\fi

\subsection{Extension to Object Detection}
We further applied our SQuantizer to object detection tasks using YOLO-v3~\cite{redmon2018yolov3} on COCO dataset~\cite{lin2014microsoft}. The dataset contains 80 classes and all the results are based on the images size of 416$\times$416 pixels.  

The backbone of YOLO-v3 is Darknet-53 and similarly as classification tasks, we did not perform SQuantizer on the first layer of the network.Training set-up are exactly the same as the baseline training with initial learning rate of 1$e^{-3}$ and divided by 10 at epoch 50. The $\sigma$ of -0.1 and -0.3 are used for 4-bit and 2-bit experiments respectively with $Delay$ at 20 epoch. The total number of training epochs is 100 epoch which is the same as the baseline training time.

\vspace{-2mm}  
\begin{table}[H]
  \center
  \caption{SQuantization on YOLO-v3 object detection with Darknet-53 on COCO Dataset with image size 416$\times$416.}
  \vspace{-2mm}  
\resizebox{\columnwidth}{!}{%
\begin{tabular}{ccccc}
\hline
 \multirow{2}{*}{} & \multirow{2}{*}{mAP}  & \multirow{2}{*}{Sparsity (\%)} & \multirow{2}{*}{\#Params (M)}& \multirow{1}{*}{Compression}\\

 \multicolumn{1}{c}{}  &  &  & &Rate  \\ \hline
 baseline & 52.8  & 0 & 61.95& -\\\hline
 %dense (4W, 32A) & 52.3  &  0 & 61.95 & 8$\times$\\\hline
 sparse (4W, 32A) & \textbf{52.0}  &  55 & 28.08 & \textbf{17$\times$}\\\hline
 sparse (2W, 32A) & \textbf{50.3}  & 46 & 33.63& \textbf{29$\times$}\\\hline
\end{tabular} 
}\vspace{-2mm}
\label{tab:yolo4bit2bit}
\end{table}

\begin{figure}[!htbp]
\vspace{-2mm}
\begin{center}
\includegraphics[height=0.3\linewidth]{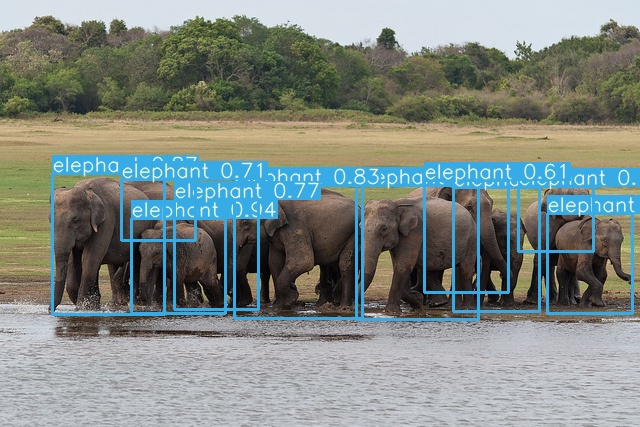}
\includegraphics[height=0.3\linewidth]{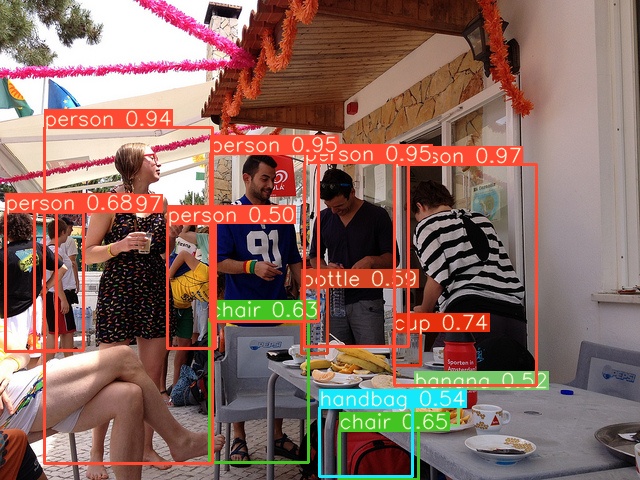}
\end{center}
\vspace{-2mm}
\caption{Output images from sparse and 4-bit quantized YOLO-v3 with Darknet-53 as a backbone classifier. Best view in color.}
\label{fig:yolo}
\vspace{-2mm}
\end{figure}

As shown in Table \ref{tab:yolo4bit2bit}, our SQuantizer boost compression rate to 17$\times$ within 1 mAP drop from the baseline and generate 55\% sparsity. With SQuantizer for 2-bit weights, it gives 29$\times$ compression rate with 50.3 mAP. Moreover, we show sample output images from the sparse and 4-bit quantized YOLO-v3 in Figure \ref{fig:yolo} for visual inspection. We believe this extension to object detection task demonstrates the good general applicability of our proposed SQuantizer.

\section{Discussion} \label{sec:discussion}
%Although both sparsification and quantization are network compression techniques. 
Mathematically, for achieving 8$\times$ compression rate, we need to either quantize a model with 4-bit precision or sparsify it with at least 87.5\% sparsity level to have an equal rate regardless of the storage overhead of non-zero elements indices. From this calculation, we can infer that low-precision (4-bit or lower) quantization can easily drive higher compression rate than sparsification. However, prior works have shown that 2-bit or lower precision quantization results in significant accuracy degradation. For example, the SOTA accuracy for ResNet50 (before our work) was 72.2\% (4.7\% drop) from \cite{choi2018pact}, when quantizing both weights and activations with 2-bit precision. Although the high (16$\times$) compression rate is attractive, the degraded accuracy may not be acceptable for real-world applications. 
This motivated us to design a joint optimization technique for sparsity and quantization, achieving maximal compression while keeping the accuracy close to the original model. As a result, we achieve 17$\times$ compression for ResNet50 when SQuantizing both weights and activations with 4-bit precision and 41\% sparsity within 1\% drop in accuracy. Table \ref{tab:flops_resnet50} shows reduction in FLOPs when leveraging weight and activation sparsity of low-bit network. %Since, in general, activations are inherently $\sim$50\% sparse due to ReLU, we expect compute reduction by half by leveraging specialized sparse hardwares ~\cite{Song2019Samsung}. 
The baseline (the first row in Table \ref{tab:flops_resnet50}) shows expected compute from existing hardwares, while the rest rows show reduced compute when leveraging the sparsity from specialized sparse hardwares ~\cite{Song2019Samsung} (i.e. 5$\times$ reduction in FLOPs from 55\% sparse 4-bit ResNet50).

%Especially for resource-constrained devices, where both high compression rate and high accuracy are important, this work can enable wider deployment of high-performing DNNs on such devices.

\vspace{-2mm}  
\begin{table}[H]
  \center
  \caption{FLOPs for dense, 4-bit vs sparse, 4-bit ResNet50}
  \vspace{-2mm}  
\resizebox{\columnwidth}{!}{%
\begin{tabular}{cccccc}
\hline
 \multirow{1}{*}{} & \multirow{1}{*}{Weights}  & \multirow{1}{*}{FLOP} & \multirow{1}{*}{Weight\%}&  \multirow{1}{*}{Activation\%}& \multirow{1}{*}{FLOP\%} \\\hline

 dense (4W, 4A) & 25.5M  & 7.7G & 100 & - & 100       \\\hline
 dense (4W, 4A) & 25.5M  & 3.2G & 100 & 40.2 & 41.2       \\\hline
 41\% sparse (4W, 4A) & 15.0M  & 1.9G &  58.8   & 42.9  & 24.6\\\hline
 55\% sparse (4W, 4A) & 11.4M  &  \textbf{1.3G} &  44.7  & 40.7  & 17.6\\\hline
\end{tabular} 
}\vspace{-2mm}
\label{tab:flops_resnet50}
\end{table}

Table\ref{tab.efficiency_1} shows potential benefits of our SQuantized model in terms of memory bandwidth, energy and computation. % and compute units. %Since our SQuantizer generates both sparse and low-bit (weight and activations) model with SOTA accuracy, we can get the maximum benefits. 
SQuantizer enables networks to fit in on-chip memory for embedded devices, which greatly reduces energy consumption compared to fetching uncompressed weights from an external DRAM. % for each image. 
%Prototyping on FPGAs shows $\sim2\times$ speedup per layer for ResNet50 with $\sim$50\% sparsity
Hardware prototyping shows $\sim$2$\times$ speedup per layer for ResNet50 with $\sim$50\% sparsity. % taking into consideration.

\vspace{-2mm}
\begin{table}[htb]
\centering
\caption{Potential benefits of compressed networks (W: weights, A: activations)}
\resizebox{1.02\columnwidth}{!}{%
\normalsize
\begin{tabular}{|l|c|c|c|}
\hline
 & Save Bandwidth? & Save Energy? & Save Compute? \\ \hline
Compress W & \checkmark & $\times$ & $\times$ \\ \hline
%Gate computation cycles for zero (W and/or A) operands & $\times$ & \checkmark & $\times$ \\ \hline
%Skip computation cycles for zero W &\checkmark  & \checkmark & \checkmark \\ \hline
Skip computation cycles for either zero W or zero A & \checkmark & \checkmark & \checkmark \\ \hline
Perform multiple lower-bitwidth operations together & \checkmark & \checkmark & $\times$ (compute units \checkmark) \\ \hline
\end{tabular}}\vspace{-3mm}
\label{tab.efficiency_1}
\end{table}%

\section{Conclusion} 
In this paper, we proposed SQuantizer, a new training method that enables simultaneous learning for sparse and low-precision (weights and activations) neural networks, provides high accuracy and high compression rate. SQuantizer outperforms across networks compared to the various prior works, even with additional sparsity. % which in turn leads to higher compression rate. 
Furthermore, the extension to YOLO-v3 for object detection demonstrates the viability and more general applicability of the proposed SQuantizer to broader vision neural network tasks. To the best of our knowledge, this paper is the first approach to demonstrating simultaneous learning for both sparse and low-precision neural networks to help the deployment of high-performing neural networks on resource-constrained edge devices for real-world applications.

{\small
\bibliographystyle{ieee}
\bibliography{egbib}
}

\end{document}